\documentclass[oribibl]{llncs}

 \pdfoutput=1
\usepackage{makeidx}  

\usepackage{graphicx}
\usepackage{amssymb,amsmath}
\usepackage{hyperref}

\usepackage{todonotes}

\begin{document}
\frontmatter          
\mainmatter              
\title{Vesselness via Multiple Scale Orientation Scores}
\titlerunning{Vesselness via Multiple Scale Orientation Scores}  
%
\author{Julius Hannink, Remco Duits and Erik Bekkers}
\authorrunning{Julius Hannink, Remco Duits and Erik Bekkers} 
%
\tocauthor{Julius Hannink, Remco Duits and Erik Bekkers}
\institute{
Eindhoven University of Technology, Department of Biomedical Engineering and Department of Mathematics and Computer Science, Eindhoven, the Netherlands\\\email{J.Hannink@gmx.de},
\email{R.Duits@tue.nl} and \email{E.J.Bekkers@tue.nl}}

\maketitle              

\hyphenation{e--stab--lished}
\begin{abstract}
The multi--scale Frangi vesselness filter is an established tool in (retinal) vascular imaging. However, it cannot properly cope with crossings or bifurcations since it only looks for elongated structures. 
Therefore, we disentangle crossings/bifurcations via (multiple scale) invertible orientation scores and apply vesselness filters in this domain. 
This new method via scale--orientation scores performs considerably better at enhancing vessels throughout crossings and bifurcations than the Frangi version. Both methods are evaluated on a public dataset. Performance is measured by comparing ground truth data to the segmentation results obtained by basic thresholding and morphological component analysis of the filtered images. 
\end{abstract}
\section{Introduction}
The retinal vasculature enables non--invasive observation of the human circulatory system. A variety of eye--related and systematic diseases such as glaucoma, age--related macular degeneration, diabetes, hypertension, arteriosclerosis or Alzheimer's disease affect the vasculature and may cause functional or geometric changes \cite{Ikram2013}. Automated quantification of these defects promises massive screenings for systematic and eye--related vascular diseases on the basis of fast and inexpensive imaging modalities, i.e. retinal photography.

To automatically assess the state of the retinal vascular tree, vessel segmentations and/or models have to be created and analyzed. Because retinal images usually suffer from low contrast at small scales, the vasculature needs to be enhanced prior to model creation/segmentation. One well--established approach is the Frangi vesselness filter \cite{Frangi1998}. It is frequently used in robust retinal vessel segmentation methods \cite{Budai2013,Lupascu2010}. However, the Frangi filter has a known drawback. It cannot properly enhance vessels throughout crossings or bifurcations that make up huge parts of the retinal vascular network. 

\begin{figure}
\parbox{.4\textwidth}{
	\centering
	\includegraphics[width=.4\textwidth]{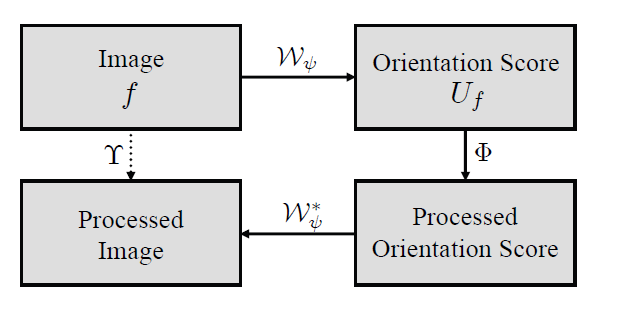}
}\hfill
\parbox{.6\textwidth}{
	\centering
	\includegraphics[width=.6\textwidth]{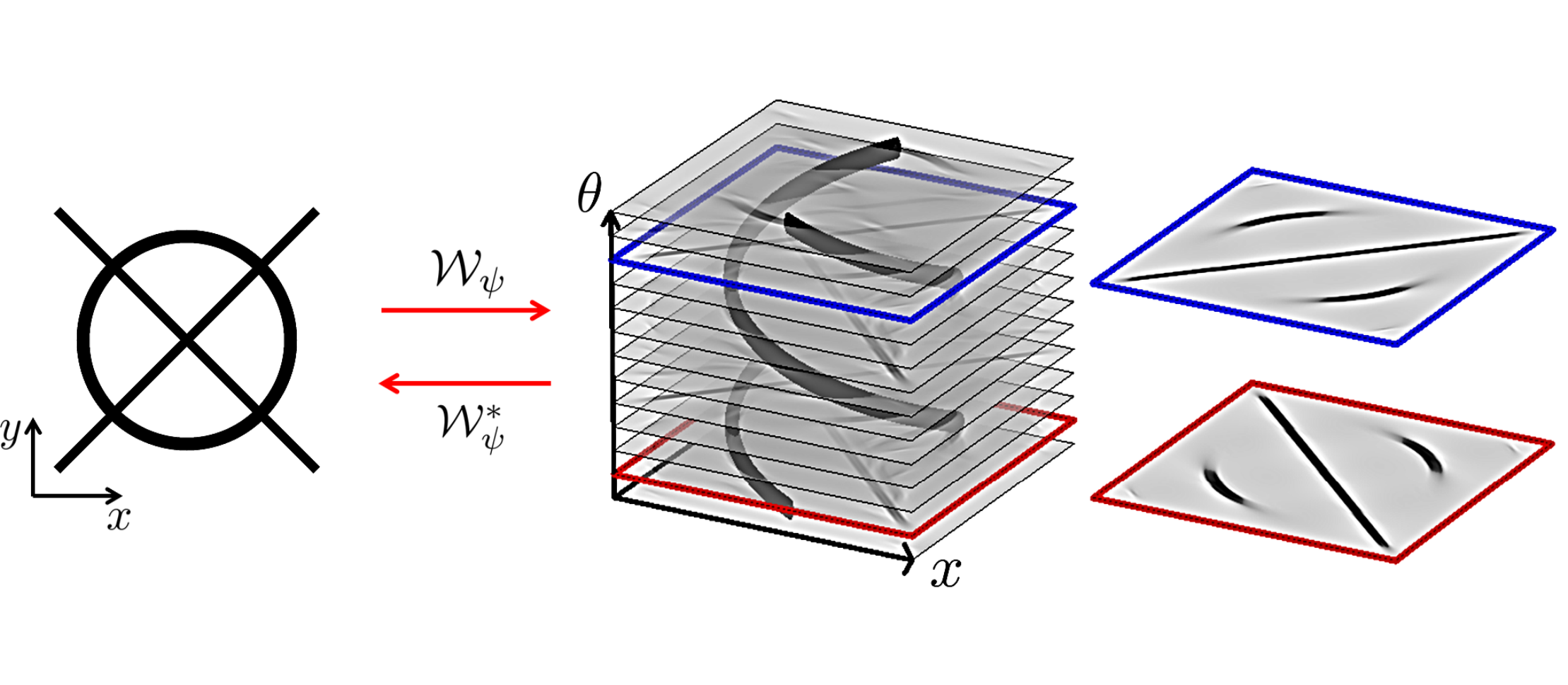}
}\hfill\null
\parbox[t]{.38\linewidth}{
	\caption{Image processing via invertible orientation scores.}
	\label{fig:ImageProcessingViaOS}
}\hfill
\parbox[t]{.58\linewidth}{
	\caption{Exemplary image and corresponding orientation score.}
	\label{fig:OSIntro}
}\hfill\null
\end{figure}

To generically deal with this issue, we apply the principle of image processing via invertible orientation scores (Fig. \ref{fig:ImageProcessingViaOS}). In the orientation score domain, crossing/bifurcating lines are disentangled into separate layers corresponding to their orientation (Fig. \ref{fig:OSIntro}). Consequently, an equivalent of the Frangi filter can be used on this domain to enhance vessels in a crossing--preserving way. 

The construction of invertible orientation scores is inspired by the functional architecture of the human cortical columns in the primary visual cortex \cite{Duits2007}, where decomposition of local orientation allows the visual system to separate crossing/bifurcating structures. Because of this, a human observer robustly identifies the two lines in the exemplary image in Fig. \ref{fig:OSIntro}, whereas conventional line--filtering on the image domain might fail in cases with noise and low contrast. 

Similar approaches of frequency or velocity specific data representations are used in other contexts \cite{Duits2012,Barbieri2013}. The general underlying theme is switching to the corresponding Lie--group of interest in the generic group theoretical approach outlined in \cite{DuitsBurgeth2007}. Here, we will develop 
vesselness filters on the extended Lie--group domains of the rotation translation group $SE(2)$ and the rotation, translation and scaling group $SIM(2)$.
Our approach is closely related to the work by Krause, Alles, Burgeth and Weickert who rely on a local Radon transform to disentangle crossings and bifurcations \cite{Krause2013}. 

The extension of vesselness filters on $\bbbr^2$ to $SE(2)$ and $SIM(2)$ allows us to cope with the high degree of (multiple scale) crossings and bifurcations in 2d retinal images. In the end, we show the performance of this new type of vesselness filters by comparison to the multi--scale Frangi vesselness \cite{Frangi1998}, both qualitatively and quantitatively on the High Resolution Fundus (HRF) image dataset available at \url{http://www5.cs.fau.de/research/data/fundus-images/}.
\null\vspace{-1.5em}\null
\null\vspace{-2.5em}\null
\section{Methods}
\subsection{Orientation Scores on $SE(2)$}
An orientation score $U_f:SE(2) \to \mathbb{C}$ is obtained by correlating an input image $f$ with a specially designed, anisotropic wavelet $\psi$:
\begin{equation}
	U_f(\vec{x},\theta) =  \left(\overline{\psi_\theta} \star f\right) (\vec{x},\theta) = \int_{\bbbr^2} \overline{\psi\left(R_\theta^{-1}(\vec{y}-\vec{x})\right)} \; f(\vec{y}) \mathrm{d}\vec{y}
	\label{eq:OSconstruction}
\end{equation}
where $R_\theta$ denotes a 2d counter--clockwise rotation matrix and $\psi_\theta(\vec x) = \psi(R_\theta^{-1}\vec x)$. 
Exact image reconstruction is achieved by
\begin{equation}
	f = \mathcal{F}^{-1}\left[M_\psi^{-1} \mathcal{F}\left[\vec{x}\mapsto \frac{1}{2\pi}\int_0^{2\pi}\left(\psi_\theta * U_f\right)(\vec{x},\theta)\mathrm{d}\theta\right]\right]
	\label{eq:OSexactreconstruction}
\end{equation}
where $*$ denotes convolution, $\mathcal{F}[\cdot]$ represents the unitary Fourier transform on $\mathbb{L}_2(\bbbr^2)$ and $M_\psi$ is given by $\int_0^{2\pi}|\mathcal{F}\left[\psi_\theta\right]|^2 \mathrm{d}\theta$.
Theoretically, reconstruction is well posed for $0<\delta<M_\psi<\infty$ with arbitrary small $\delta$. Practically, however, it has proven best to aim at $M_\psi\approx 1$ since that ensures optimal stability \cite{Fuehrbook}.

One type of wavelets that meet this stability criterion are the cake wavelets described by \cite{FrankenPHD2008,Bekkers2014}. They uniformly cover the Fourier domain up to a radius of about the Nyquist frequency $\rho_n$ to satisfy the discrete version of $M_\psi \approx 1$ by design. This is also the general idea behind curvelets and shearlets \cite{Candes2006,Bodmann2013}. They do, however, not give appropriate representations in our context. Curvelets have varying orientation localization over scale, whereas retinal images show detail on all scales. Shearlets are based on the shearing, translation and scaling group, whereas rotation invariance is much more desirable in our context. 

\begin{figure}[!h]
		\centering
		\includegraphics[width=\textwidth]{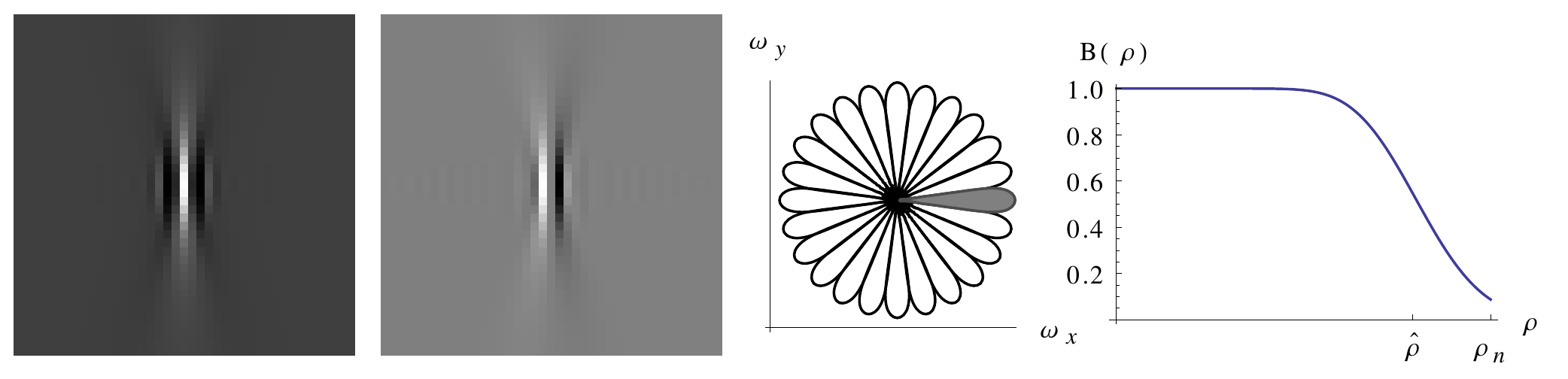}
		\caption{Real and imaginary part of the cake kernel (left, zoomed), Fourier contours at 70\% of the maximum for all orientations and $B(\rho)$ with $\hat\rho=0.8\rho_n$, the Nyquist frequency $\rho_n$ and $N=8$.}
		\label{fig:cakewavelets}\vspace{-1em}
\end{figure}

The procedure for creating cake wavelets is illustrated in Fig. \ref{fig:cakewavelets}
. In detail, the proposed wavelet takes the form
\begin{equation}
	\psi(\vec{x}) = \left(\mathcal{F}^{-1}\left[\vec{\omega}\mapsto\tilde\psi(\rho \cos \varphi,\rho \sin\varphi)\right]\right)(\vec x)\; G_s(\vec{x})
	\label{eq:cakewavelets}
\end{equation}
where $G_s$ is an isotropic Gaussian window in the spatial domain and $(\rho,\varphi)^T$ denote polar coordinates in the Fourier domain, i.e. $\vec\omega = (\rho \cos\varphi,\rho\sin\varphi)^T$. The Fourier wavelet $\tilde\psi(\vec \omega) = A(\varphi)B(\rho)$ is constructed from
\begin{equation}
\begin{array}{lr}
A(\varphi) = \left\{
	\begin{array}{ll}
		B^k\left(\frac{(\varphi\, \mathrm{mod}\, 2\pi)-\pi/2}{s_\theta}\right)& \mathrm{if}\;\rho > 0\\
		\frac{1}{N_\theta} &\mathrm{if}\;\rho = 0
	\end{array}
\right.&\hspace{1em},\hspace{1em}
B(\rho) = e^{-\left(\frac{\rho}{t}\right)^2} \sum\limits_{i=0}^N \frac{(\rho/t)^{2i}}{i!}\\
\end{array}
\label{eq:cakewaveletsFourier}
\end{equation}
$B^k(x)$ denotes the $k-$th order B--spline, $N_\theta$ is the number of samples in the orientation direction and $s_\theta=2\pi/N_\theta$ is the angular stepsize. The function $B(\rho)$ is a Gaussian multiplied with the Taylor series of its inverse up to order $N$ to enforce faster decay. The parameter $t$ is given by $t^2=2 \hat\rho/(1+2N)$ with the inflection point $\hat\rho$ that determines the bending point of $B(\rho)$.

As depicted in Fig. \ref{fig:cakewavelets}, the real part of the kernel picks up lines, whereas the imaginary part responds to edges. In the context of vessel filtering, the real part is of primary interest, whereas vessel tracking in the score additionally makes use of the imaginary part \cite{Bekkers2014}. 

\subsection{Gaussian Derivatives in Orientation Scores}
The orientation score domain is essentially the 2d special Euclidean motion group $SE(2)\equiv \bbbr^{2} \rtimes S^{1}$ \cite[ch.2]{FrankenPHD2008}. Because of this, all operations $\Phi$ on this domain have to be left--invariant to produce a Euclidean invariant net operator $\Upsilon$ on the image \cite[ch.2]{FrankenPHD2008}. This is desirable since the result should be independent on rotation and translation of the input. 
$\Phi$ is left--invariant iff $\Phi \circ \mathcal{L}_{\vec g} = \mathcal{L}_{\vec g} \circ \Phi$ for all $\vec g=(\vec x,\theta) \in SE(2)$ with the shift--twist operator on the score given by 
\begin{equation}
	\mathcal{L}_{\vec g}U_f(\vec h)= U_{f}(\vec g^{-1}\vec h)= U_{f}(R_{\theta}^{-1}(\vec{x}'-\vec{x}),\theta'-\theta),
\end{equation}
for all $\vec g=(\vec x,\theta), \vec h=(\vec x',\theta') \in SE(2)$. Therefore, we must rely on left--invariant derivatives, given by $ \partial_{\theta}, \partial_{\xi}= \cos \theta \partial_{x} +\sin \theta \partial_{y}$ and $ \partial_{\eta}= -\sin \theta \partial_{x} + \cos \theta \partial_{y}$ when constructing vesselness filters on $SE(2)$. These derivatives provide a moving frame of reference on the group steered by the orientation of the wavelet. Their non--zero commutators are given by $[\partial_\theta,\partial_\xi]=\partial_\eta$ and $[\partial_\theta,\partial_\eta]=-\partial_\xi$.
Later we will adapt this frame 
locally to the score, following the theory of best exponential curve fits presented in \cite[ch.6]{FrankenPHD2008}. This compensates for the fact that our wavelet kernel is not always perfectly aligned with all local orientations present in the image (For details, see \cite[ch.6]{FrankenPHD2008}).

\begin{figure}
\null\hfill
\parbox{.5\textwidth}{
	\centering
	\includegraphics[width=.46\textwidth]{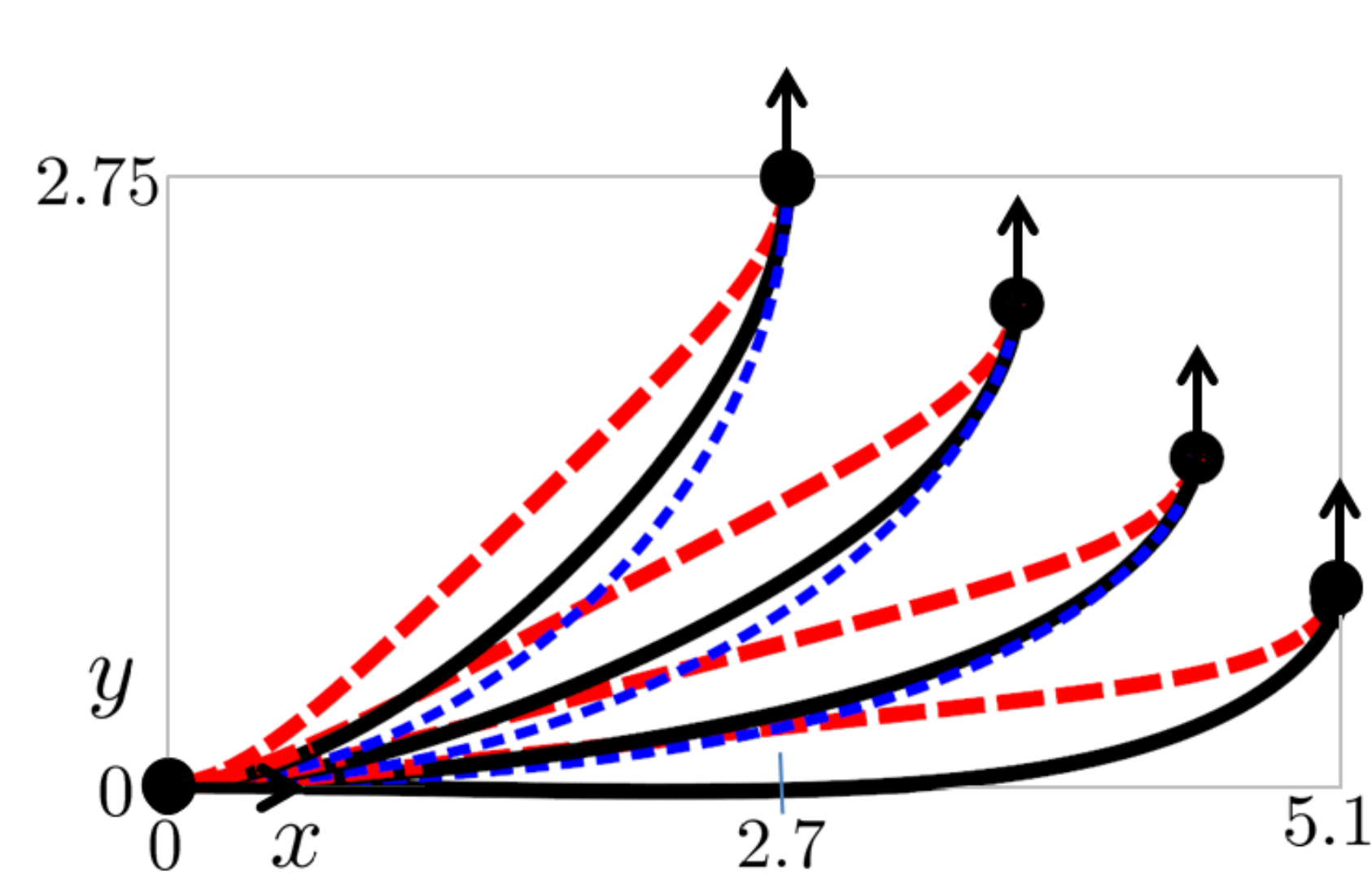}\hfill\null
}\hfill
\parbox{.48\textwidth}{
	\centering
	\includegraphics[width=.46\textwidth]{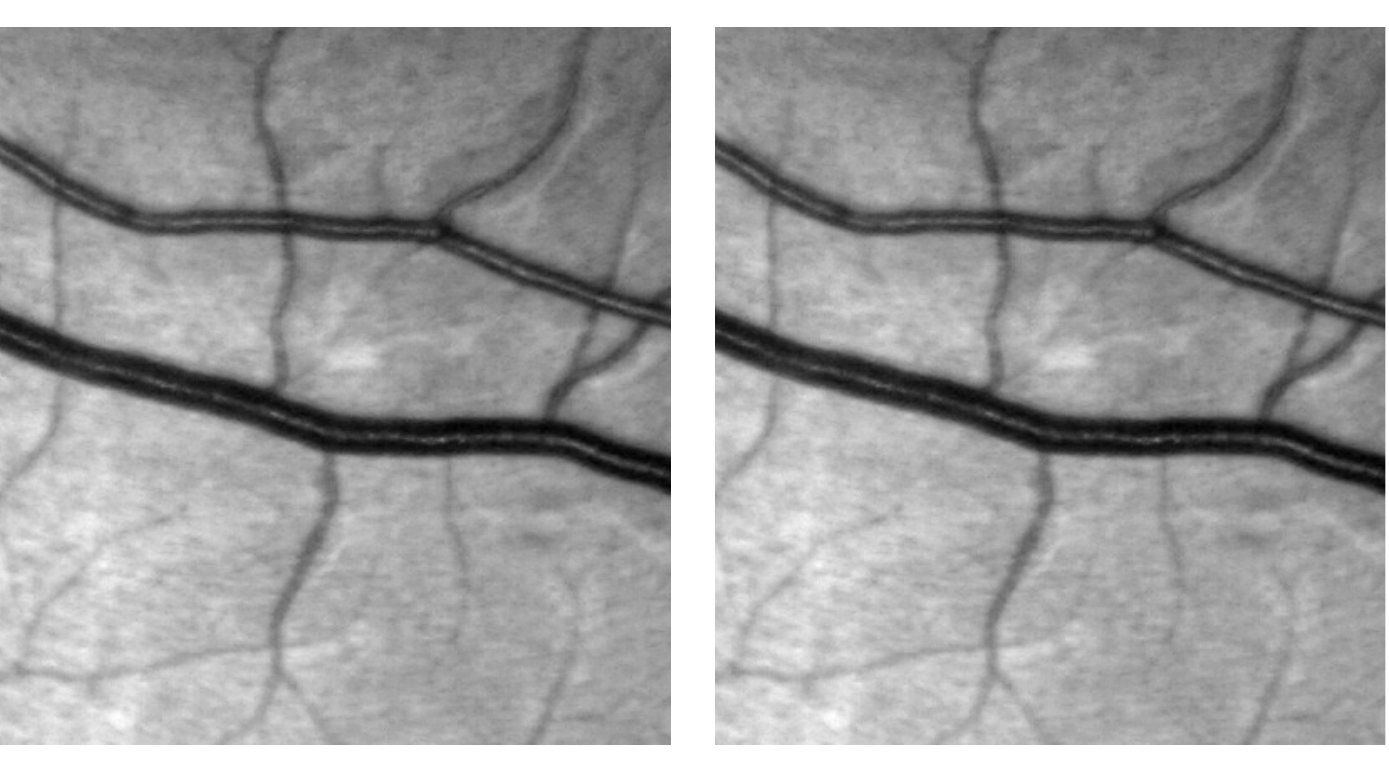}\hfill\null\vspace{-1em}
}\hfill\null
\null\hfill
\parbox[t]{.48\linewidth}{
		\caption{Spatial projection of $SE(2)$ geodesics for $\beta=3$ (dashed, red), $\beta=1$ (solid, black) and $\beta=1/3$ (thin, dashed, blue). For details on sub--Riemannian geodesics and association fields, see \cite{Duits2013}.}%
		\label{fig:geodesicsSE2}
}\hfill
\parbox[t]{.45\linewidth}{
	\caption{Exemplary retinal image (left) and reconstruction from its scale--orientation score (right).}%
	\label{fig:MSReconstruction}
}\hfill\null
\end{figure}

Since orientation and spatial direction have different physical units, a conversion factor is needed. This parameter $\beta$ has unit 1/length and determines the shape of $SE(2)$ geodesics (see Fig. \ref{fig:geodesicsSE2}). Mathematically, $\beta$ appears as the only free parameter in 
the (sub-)Riemannian metric on $SE(2)$ given by
\begin{equation}
d(\vec g_1,\vec g_2) = \inf_{{{\gamma(0)=\vec g_1}\atop{\gamma(l)=\vec g_2}}\atop{\dot\gamma\in\Delta, l\geq 0}} \int_0^l \sqrt{\mathcal{G}_\beta\big|_{\gamma(s)}\big(\dot\gamma(s),\dot\gamma(s)\big)}\; \mathrm{d}s\;\;,
\label{eq:metrictensor}
\end{equation}
with $\Delta=\mbox{span}\,\{\partial_\xi,\partial_\eta,\partial_\theta\}$, $\gamma(s)=(\vec x(s),\theta(s))$ and $\mathcal{G}_\beta|_\gamma(\dot\gamma,\dot\gamma) =  \beta^2(\dot x \cos\theta + \dot y \sin\theta)^2 + \beta^2(-\dot x \sin\theta +  \dot y \cos\theta)^2 + \dot\theta^2$ in the Riemannian case. 
In the sub--Riemannian case, the allowed part of the tangent space is $\Delta=\mbox{span}\,\{\partial_\xi,\partial_\theta\}$. The functional in (\ref{eq:metrictensor}) then reduces to $\int_0^l\sqrt{\kappa^2(s)+\beta^2}\;\mbox{d}s$ with the curvature $\kappa$ of the spatially projected curve $\vec x(s)=\mathcal{P}_{\bbbr^2}\gamma(s)$ under the condition that $\vec g_2$ is chosen ``aligned enough''  with $\vec g_1$ (For details, see \cite{Duits2013}). Typically, $\beta<1$.

In order to extract local features in $SE(2)$ with well--posed, left--invariant derivative operators, some regularization has to be included. 
The only left--invariant diffusion regularization in $SE(2)$ that preserves the non--commutative group structure via the commutators is elliptic diffusion, isotropic w.r.t. the $\beta$-metric (\ref{eq:metrictensor}). In this case, regularization is achieved via a spatially isotropic Gaussian with scale $\frac{1}{2}\sigma_{s}^{2}$ and a 1d--Gaussian in $\theta$ with scale $\frac{1}{2}(\beta\sigma_s)^{2}$ \cite[ch.5]{FrankenPHD2008}.
The regularized derivative operators are convolutions with correspondingly differentiated $\beta$-isotropic Gaussians and generalize the concept of Gaussian derivatives used in the Frangi vesselness filter \cite{Frangi1998} to $SE(2)$. 

\subsection{Scale--Orientation Scores}
\begin{figure}
		\centering
		\includegraphics[width=\textwidth]{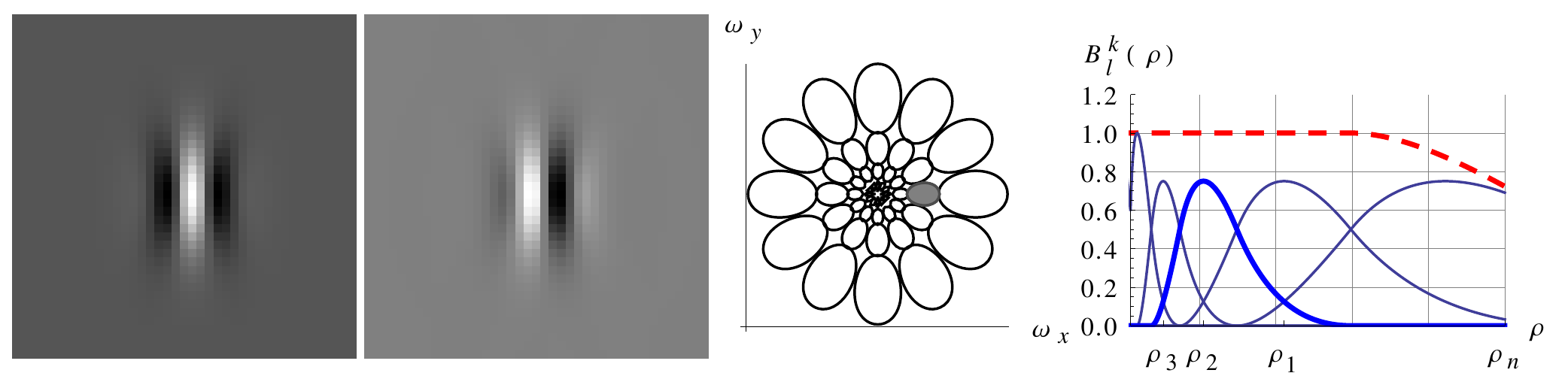}\vspace{-1em}
		\caption{Real and imaginary part of the multi--scale cake kernel at $a_2$ (left), Fourier contours of all wavelets (at 70\% of maximum) and $B^\mathrm{MS}(\rho)$ (red, dashed) with components $B^k_l(\rho)$ (blue). 
		Highlighted components correspond to the kernels shown.}
		\label{fig:MScakewavelets}
\end{figure}

To make the kernels described above scale--selective, the pieces of cake have to be further divided. By cutting out pieces in the log--radial direction, they are made sensitive to a specific frequency range that can be identified with a scale $a$ in the spatial domain. 
To construct scale selective cake kernels (anisotropic wavelets), \cite{Sharma2013} uses a radial envelope function
\begin{equation}
	B^\mathrm{MS}(\rho)=\sum\limits_{l=0}^{{N_\rho}-1}B_l^k(\rho) := \sum_{l=0}^{{N_\rho}-1}B^k\left(\frac{\log(\rho a^-)}{s_\rho}+l\right)
	\label{eq:SOSradialenvelope}
\end{equation}
where $B^k(x)$ is the $k$-th order B--spline function, $N_\rho$ is the total number of scales to sample in the Fourier domain and $s_\rho>0$ denotes the stepsize in log--scale. The multiplicative character of the spatial scales $a_l=a^{-} e^{l s_{\rho}}$ reflects the typical scale transition at bifurcations. Because of the B--spline approach, the scale selective envelopes $B^k_l(\rho)$ sum  to one and the $M_\psi\approx 1$ requirement is still met (Fig.\ref{fig:MScakewavelets}).
Scale layers outside a spatially defined range of interest are merged to reduce computational load (Fig. \ref{fig:MScakewavelets}).
We propose the following multi--scale cake kernel
\begin{equation}
	\psi^{\mathrm{MS}}(\vec{x}) = \left(\mathcal{F}^{-1}\left[M^{-1}\mathcal{F}\left[\tilde\psi^\mathrm{MS}\right](\boldsymbol\omega)\right]\right)(\vec x)
	\label{eq:MSCakeKernels}
\end{equation}
where $\tilde\psi^\mathrm{MS}(\vec{x})$ denotes the wavelet 
\begin{equation}
	\tilde\psi^{\mathrm{MS}}(\vec{x}) = \left(\mathcal{F}^{-1}\left[\vec\omega\mapsto A(\varphi)B_0^k(\rho)\right]\right)(\vec x)\;G_{s_x,s_y}(\vec x)
	\label{eq:MSCakeKernels1}
\end{equation}
at the finest scale $a^-$. The anisotropic Gaussian window $G_{s_x,s_y}(\vec x)$ reduces long tails along the orientation of the wavelet and suppresses oscillations perpendicular to it induced by narrow sampling bandwidths in $B^\textrm{MS}(\rho)$. The corresponding changes in the Fourier domain are resolved by subsequent normalization with $M(\omega)= N_\rho^{-1}N_\theta^{-1} \sum_{i=1}^{N_\rho}\allowbreak \sum_{j=1}^{N_{\theta}} a_i^{-1}\allowbreak|\mathcal{F}[\tilde{\psi}^{MS}]\allowbreak(a_{i}R_{\theta_{j}}^{-1} \vec\omega)| $. Thereby, reconstruction is done by summation over scales and angles and perfect invertibility (Fig. \ref{fig:MSReconstruction}) is ensured. The scale and orientation specific data representation given by $(\mathcal{W}_\psi f)(\vec x,\theta,a) = (\overline{\psi_\theta^a} \star f)(\vec x)$ with $\psi_\theta^a(\vec x) =a^{-1} \psi^\mathrm{MS}(a^{-1} R_\theta^{-1} \vec x)$ is now set and processing can begin.

\null\vspace{-2.5em}\null
\subsection{Vesselness Filtering on Scale--Orientation Scores}
The single--scale Frangi vesselness filter $\mathcal{V}_0^\mathrm{Fr}$ makes use of an anisotropy measure $\mathcal{R}$ and a structure measure $\mathcal{S}$ based on second order derivatives in a coordinate system aligned with elongated structures \cite{Frangi1998}. This approach is now generalized to (scale--)orientation scores for crossing--preserving vesselness filtering. 
Given a convexity criterion $\mathcal{Q}>0$ on transversal vessel profiles and the measures $\mathcal{R}$ resp. $\mathcal{S}$, the $SE(2)$ vesselness expression $\mathcal{V}_0^{SE(2)}(\mathcal{U}_f^a): SE(2) \to\bbbr^+$  is identical to the one proposed by \cite{Frangi1998}:
\begin{equation}
	\mathcal{V}_0^ {SE(2)}\, (\mathcal{U}_f^a) = \left\{
	\begin{array}{ll}
	0 & \mbox{ if }\mathcal{Q} \leq 0 \\
	\exp\left(-\frac{\mathcal{R}^2}{2 \sigma_1^2}\right)\left[1-\exp\left(-\frac{\mathcal{S}}{2\sigma_2^2}\right)\right]	& \mbox{ if } \mathcal{Q}> 0
	\end{array}
	 \right.
	\label{eq:FrangiSE2}
\end{equation}
where $\mathcal{U}_f^a (\vec x,\theta) = (\mathcal{W}_\psi f)(\vec x,\theta,a),\; a>0$ fixed, is a single scale layer of the multi--scale  wavelet transform. Typically, $\sigma_1=0.5$ and $\sigma_2=0.2\;||\mathcal{S}||_\infty$.

There are two natural generalizations of $\mathcal{V}_0^\mathrm{Fr}$ to $SE(2)$ that differ in the choice of coordinate system used to define $\mathcal{R},\mathcal{S}$ and $\mathcal{Q}$. One option is to work in the moving frame of reference $\{\partial_\xi,\partial_\eta,\partial_\theta\}$.  Since $\xi$ and $\eta$ are respectively parallel and orthogonal to the orientation in each $\theta$--slice, second order Gaussian derivatives along these directions carry the same information as the eigenvalues of the Hessian used in $\mathcal{V}_0^\mathrm{Fr}$ \cite{Frangi1998}. Therefore, $\mathcal{R},\mathcal{S}$ and $\mathcal{Q}$ are computed as
\begin{equation}
	\mathcal{R}=\frac{(\partial_\xi^2\,\mathcal{U}_f^a)^{s,\beta}}{(\partial_\eta^2\,\mathcal{U}_f^a)^{s,\beta}}\;\;\;, \;\;\;
	\mathcal{S}=\left[(\partial_\xi^2\,\mathcal{U}_f^a)^{s,\beta}\right]^2+\left[(\partial_\eta^2\,\mathcal{U}_f^a)^{s,\beta}\right]^2\;\;\;, \;\;\;
	\mathcal{Q}=(\partial_\eta^2\,\mathcal{U}_f^a)^{s,\beta}
	\label{eq:measuresVSE2_moving}
\end{equation}
where the superscripts $^{s,\beta}$ indicate Gaussian derivatives at spatial scale $s=\frac{1}{2}\sigma_s^2$ and angular scale $\frac{1}{2}(\beta\sigma_s)^2$.
The generalization of the filter in the $\{\partial_\xi,\partial_\eta,\partial_\theta\}$ frame is referred to as $\mathcal{V}_0^{\xi,\eta}$ at single scales and as $\mathcal{V}^{\xi,\eta}$ in the multiple scale recombination, similar to the notation in \cite{Frangi1998}.
The other possible coordinate system is the Gauge frame $\{\partial_{\vec{a}},\partial_{\vec{b}},\partial_{\vec{c}}\}$ determined by the eigendirections of the Hessian $(\mathcal{H}^{s,\beta}\,\mathcal{U}_f^a)(\vec g)$ at scale $s$ and $\vec{g}\in SE(2)$, normalized w.r.t the $\beta$--metric (\ref{eq:metrictensor}). As the filter is no longer confined to $\theta$-slices because the Gauge frame is free to fully align with the data in the score, the analogies to \cite{Frangi1998} are even stronger in this frame. Given the eigenvalues of the Hessian $(\mathcal{H}^{s,\beta}\,\mathcal{U}_f^a)(\vec g)$ ordered in absolute magnitude $|\lambda_1|\leq|\lambda_2|\leq|\lambda_3|$, $\mathcal{R},\mathcal{S}$ and $\mathcal{Q}$ are computed as
\begin{equation}
	\mathcal{R}=\frac{\lambda_1}{c} \;\;\;, \;\;\;
	\mathcal{S}=\lambda_1^2 + c^2  \;\;\;, \;\;\;
	\mathcal{Q}=c
	\label{eq:measuresVSE2_abc}
\end{equation}
with $c=\frac{1}{2}(\lambda_2+\lambda_3)$. As such, $c$ is comparable to the orientation confidence defined by \cite{FrankenPHD2008}. The generalization of the vesselness filter in this frame is referred to as $\mathcal{V}_0^{\vec{a},\vec b, \vec c}$ at single scales and as $\mathcal{V}^{\vec{a},\vec b, \vec c}$ in the multi--scale recombination.

The generalization of the multi--scale Frangi filter $\mathcal{V}^\mathrm{Fr}$ to the $SIM(2)$ domain is achieved by image reconstruction from vesselness filtered scale--orientation scores with subsequent intensity normalization 
\begin{equation}
	\big(\mathcal{V}^{SIM(2)} (f)\big)(\vec x)= \mu_\infty^{-1} \sum \limits_{i=1}^{N_{s}} \mu_{i,\infty}^{-1} \sum \limits_{j=1}^{N_{\theta}} \big(\mathcal{V}_{0}^{SE(2)}(U_{f}^{a_{i}})\big)(\vec x,\theta_{j})
	\label{eq:vesselnessSIM2}
\end{equation}
where $\mu_\infty$ and $\mu_{i,\infty}$ are the maximum values, i.e. $||\cdot||_\infty$ norms, taken over the subsequent sums. To get comparable results, the multi--scale Frangi vesselness filter is also computed via summation over single scale results and normalization by the maximum value.

\begin{figure}
		\centering\vspace{-.5em}
		\includegraphics[width=\textwidth]{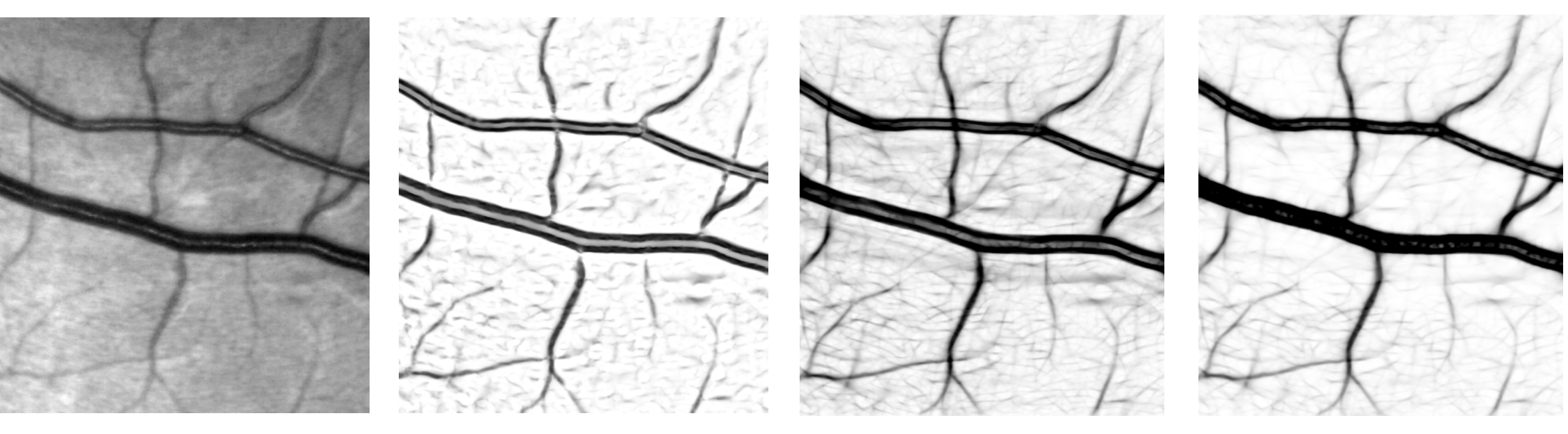}\vspace{-1em}
		\caption{
		Retinal image $f$ and multi--scale vesselness filtering results for the Frangi filter $\mathcal{V}^\mathrm{Fr}(f)$ and our two methods $\mathcal{V}^{\xi\eta}(f)$ resp. $\mathcal{V}^{\vec a, \vec b,\vec c}(f)$ (left to right). Gray scales adjusted.}
		\label{fig:vesselnessosmethodsMS}
		\null\vspace{-1.5em}\null
\end{figure}

Fig. \ref{fig:vesselnessosmethodsMS} shows multi--scale vesselness results on an exemplary retinal image obtained with the Frangi filter and our two methods for five scales $\{1.5, 2.4, 3.8, \allowbreak6.0, 9.5\}$ px, $\beta=0.05/a$ and 12 orientations sampled in $[0,\pi)$.
 Both our methods clearly outperform the Frangi filter at crossings and bifurcations. The Gauge--frame method $\mathcal{V}^{\vec a,\vec b,\vec c}$ gives best results since it is better aligned with elongated structures in the score. 
\newpage
\section{Experiments}
To show the benefit of crossing--preservation in multiple scale vesselness filtering, we devised a simple segmentation algorithm to turn a vesselness filtered image $\mathcal{V}(f)$ into a binary vessel map. First, an adaptive thresholding is applied, yielding a binary image
\begin{equation}
f_B = \Theta\big([\mathcal{V}(f)-G_\gamma*\mathcal{V}(f)]-t\big)
\label{eq:IB}
\end{equation}
where $\Theta$ is the Heaviside step function  and $G_\gamma$ is a Gaussian of scale $\gamma\gg1$. In a second step, the connected morphological components in $f_B$ are subject to size and elongation constraints. Components counting less than $\tau$ pixels or showing elongations below a threshold $\nu$ are removed. The parameters $\gamma, \tau$ and $\nu$ are fixed at 100 px, 500 px and 0.85 respectively. $\mathcal{V}(f)$ is either obtained via $\mathcal{V}^\mathrm{Fr}$ or via the $SIM(2)$ method $\mathcal{V}^{\vec a, \vec b,\vec c}$ with the settings mentioned earlier.
 
\begin{figure}
		\centering\vspace{-0.7em}
		\includegraphics[width=\textwidth]{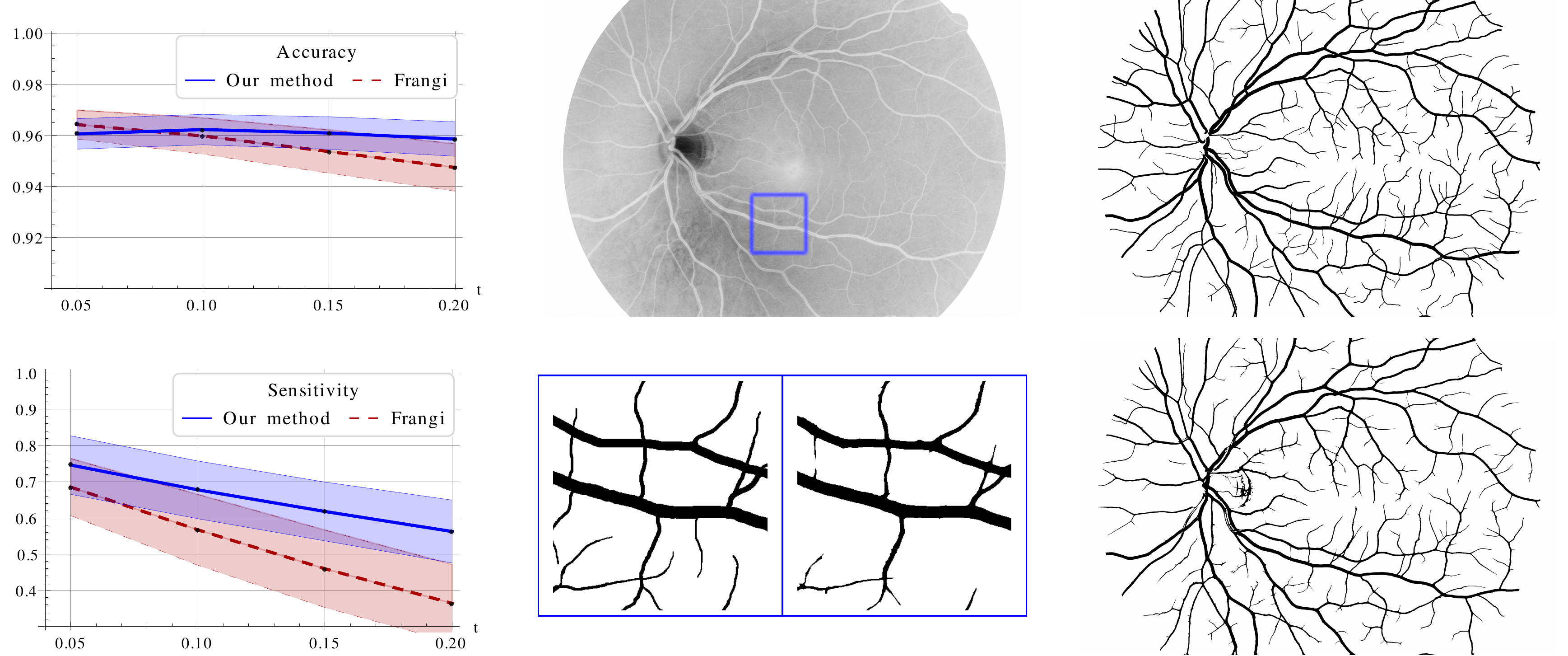}\vspace{-.5em}
		\caption{Average accuracy and sensitivity on the HRF dataset over threshold values $t$. Shaded regions correspond to $\pm1 \,\sigma$ (left). Retinal image and patch--wise ground truth/segmentation at $t=0.05$ (center). Full ground truth and segmentation (right).}
		\label{fig:experiments} 
		\null\vspace{-1.5em}\null
\end{figure}

This segmentation algorithm is evaluated on the HRF dataset consisting of wide--field fundus images for a healthy, diabetic retinopathy and glaucoma group (15 images each, ground truths provided). Average sensitivity and accuracy on the whole dataset are shown in Fig. \ref{fig:experiments} over threshold values $t$. Our method via invertible scale--orientation scores performs considerably better than the method based on the multi--scale Frangi filter. The segmentation results obtained with $\mathcal{V}^{\vec a,\vec b,\vec c}$ are more stable w.r.t variations in the threshold $t$ and the performance on the small vasculature has improved as measured via the sensitivity. Average sensitivity, specificity and accuracy at a threshold $t=0.05$  resp. given by $0.786,0.988,0.969$ (healthy), $0.811,0.963,0.953$ (diabetic retinopathy) and $0.641,0.988,0.960$ (glaucoma) compare well with other algorithms evaluated on the HRF dataset (see \cite[Tab. 5]{Budai2013}). On the diabetic retinopathy group, our method even outperforms existing segmentation methods.
Fig. \ref{fig:experiments} shows a full segmentation computed with the proposed method and an in--detail patch.
\null\vspace{-2.5em}\null
\section{Discussion}
We developed (multi--scale) crossing--preserving vesselness filters as generalizations of \cite{Frangi1998} to the extended Lie group domains $SE(2)$ resp. $SIM(2)$. The new filters were evaluated qualitatively and quantitatively on a public dataset and outperformed the Frangi filter and existing segmentation methods. This shows the method's potential for application in other areas of vascular imaging. Reduced sensitivity on the glaucoma group is most likely due to non--uniform contrast in the input. Future work therefore includes contrast normalization as preprocessing and concatenation with enhancements \cite{FrankenPHD2008, Sharma2013} and tracking \cite{Bekkers2014}.

\vspace{1em}
{\scriptsize\textbf{Acknowledgements}: The research leading to these results has received funding from the ERC councilunder the EC's 7th Framework Programme (FP7/2007--2013) / ERC grant agr. No. 335555.}
\vspace{-1em}
%
%
\bibliographystyle{splncs}
\bibliography{References}
\end{document}